\documentclass{WileyMSP-template}

\usepackage{graphicx}
\usepackage{epstopdf}
\usepackage{ragged2e}
\usepackage{amsmath, amsfonts, amssymb, amsthm}
\usepackage{cite}
\usepackage{float}
\usepackage{lineno}
\usepackage{xcolor}

\begin{document}

\pagestyle{fancy}
\rhead{\includegraphics[width=2.5cm]{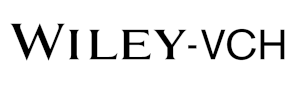}}

\title{Proprioceptive and Exteroceptive Information Perception in a\\ Fabric Soft Robotic Arm via Physical Reservoir Computing \textcolor{red}{with minimal training data}}
\maketitle

\author{Jun Wang$^1$*, Zhi Qiao$^2$, Wenlong Zhang$^3$, and Suyi Li$^1$}

\begin{affiliations}

1. Department of Mechanical Engineering, Virginia Tech, Blacksburg, VA USA\\

\medskip
2. School for Engineering of Matter, Transport and Energy, Arizona State University, 
Tempe, AZ USA\\

\medskip
3. School of Manufacturing Systems and Networks, Arizona State University
Mesa, AZ USA\\

\medskip
*Correspondent email address: \texttt{junw@vt.edu}

\end{affiliations}


\begin{abstract}
\textbf{Abstract}

Over the past decades, we have witnessed a rapid emergence of soft and reconfigurable robots thanks to their capability to interact safely with humans and adapt to complex environments. However, their softness makes accurate control very challenging. High-fidelity sensing is critical in improving control performance, especially posture and contact estimation. To this end, traditional camera-based sensors and load cells have limited portability and accuracy, and they will inevitably increase the robot's cost and weight. In this study, instead of using specialized sensors, we only collect distributed pressure data inside a pneumatics-driven soft arm and apply the physical reservoir computing principle to simultaneously predict its kinematic posture (i.e., bending angle) and payload status (i.e., payload mass). Our results show that, with careful readout training, one can obtain accurate bending angle and payload mass predictions via simple, weighted linear summations of pressure readings. In addition, our comparative analysis shows that, to guarantee low prediction errors within 10\%, bending angle prediction requires less training data than payload prediction. This result reveals that balanced linear and nonlinear body dynamics are critical for the physical reservoir to accomplish complex proprioceptive and exteroceptive information perception tasks. Finally, the method of exploring the most efficient readout training methods presented in this paper could be extended to other soft robotic systems to maximize their perception capabilities.

\end{abstract}

\section{Introduction}

\justifying

Over the past decades, we have witnessed a rapid emergence of soft and reconfigurable robots because of their capability to interact safely with humans and operate in complex environments \cite{laschi2016soft, polygerinos2017soft, yumbla2021human}. It is also widely recognized that materializing the full potential of soft robots requires multi-faceted efforts, ranging from their design and actuation all the way to control and autonomy. In this regard, there has been a proliferation of studies in soft robotic design, fabrication, and actuation mechanisms \cite{pinskier2022bioinspiration, zaidi2021actuation}, with several of them being successfully commercialized in healthcare and manufacturing. On the other hand, progress in the sensing and control of soft robots is relatively nascent \cite{della2023model, wang2022control}. In particular, integrated and reliable sensing -- the pre-requisite of effective control and autonomy -- remains an open challenge \cite{hegde2023sensing}. 

Soft robots typically exhibit complex deformations internally and interact with the external environment, so the sensors embedded in their body need to provide two types of information. One is \textit{proprioception}, related to the robotic body’s movement, shape, and location, and the other is \textit{exteroception}, related to external stimulation such as touch, pressure, and temperature \cite{faris2023proprioception, shu2023machine}. Correspondingly, a variety of soft sensors has been developed based on different working principles.  For example, resistive (and piezo-resistive) sensors can directly measure the local strain of a soft actuator \cite{dickey2008eutectic, wang2018self}, capacitive sensors can sense the physical touch on the robotic skin \cite{atalay2017batch}, and flexible optical fiber sensors can be distributed throughout the robotic body to inform the overall shape changes \cite{larson2016highly}. Finally, different kinds of sensors can be integrated to obtain a comprehensive and accurate understanding of the robots' working conditions (aka. multi-modal sensing or sensor fusion \cite{kim2021probabilistic, loo2022robust, dai2021flexible}). While the readings from these sensors are quite simple, such as resistance, pressure, and light intensity, they can be sent to a centralized processor to extract more complex information like body kinematics or the target object's stiffness.  This ``information perception'' is sometimes accomplished with the help of data-driven and machine learning methods \cite{scimeca2019model}, and the result can inform control and decision-making.

However, adding soft sensors can complicate the overall robot design and fabrication setup, resulting in a significant increase in the robot's cost and, sometimes, a decrease in mechanical flexibility. A large array of sensors could also produce substantial data, demanding high computational power and creating time delays. But most importantly, an over-reliance on these additional and complex sensors can lead us to neglect an important source of sensing information: \textit{the soft robotic body itself.}

\begin{figure}[t]
    \centering
    \includegraphics[scale=1.0]{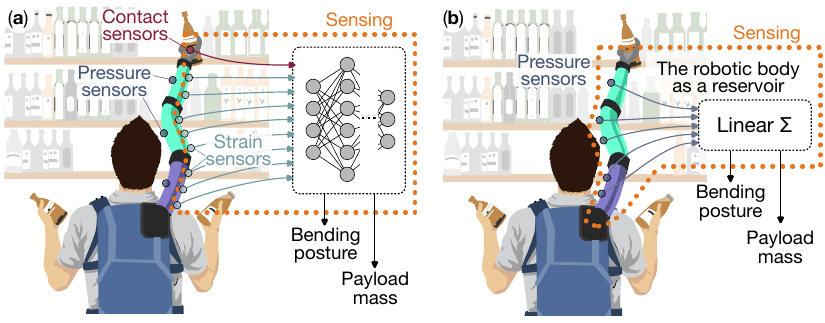}
    \caption{A ``big-picture'' comparison of two different approaches for pneumatic soft robot sensing (a) One can embed different kinds of sensors into the soft robotic body and process their readings --- using machine learning, for example --- to obtain proprioceptive and exteroceptive information. In this case, pressure sensors are typically used for actuation control only. (b) One can also treat the robotic body as a reservoir computer (i.e., a ``physical neural network'') and obtain complex information by simple, weighted linear summations of its pressure sensor readings. The soft robotic arm figure is adopted from~\cite{nguyen2019soft}.}
    \label{fig: vision}
\end{figure}

As a soft robot is activated and interacts with the environment, its body's deformation and dynamic responses inherently harbor many useful and complex proprioceptive and exteroceptive information \cite{nakajima2013soft} --- That is, the soft robotic body itself can be part of the ``sensory system'' with embodied information. Therefore, one can potentially use only a few simple sensors (preferably those ``already there'') to measure the body dynamics and then process the readings to acquire more complex and useful information.  

To this end, employing the framework of physical reservoir computing is a promising strategy to achieve such ``information perception through body mechanics.''  Reservoir computing is a branch within the discipline of artificial neural networks. In this architecture, the interconnection weights inside the neural network’s kernel remain fixed, and only readout weights are trained to reach the targeted outputs \cite{tanaka2019recent, nakajima2020physical, pieters2022leveraging, shougat2021hopf, tsunegi2019physical, bhovad2021physical, inubushi2017reservoir}.  Since the neural network does not change during training, one can use a physical body as the reservoir and harvest its nonlinear and high-dimensional dynamic responses as the computational resource (essentially, the physical body becomes the neural network; see Fig. \ref{fig: vision}).  Simple mechatronic components like embedded sensors are required to construct a physical reservoir, but computing occurs entirely in the mechanical domain.  Offloading portions of computing burdens to the mechanical domain brings significant benefits like lower power consumption, fewer digital-analog conversions, and much better reliability against difficult working conditions. \emph{Most importantly, the physical reservoir framework provides a gateway to access the sensory information embodied in the mechanical body} \cite{wang2023building, kawase2021pneumatic}. 

\begin{figure}[t!]
   \centering
    \includegraphics[scale=1.0]{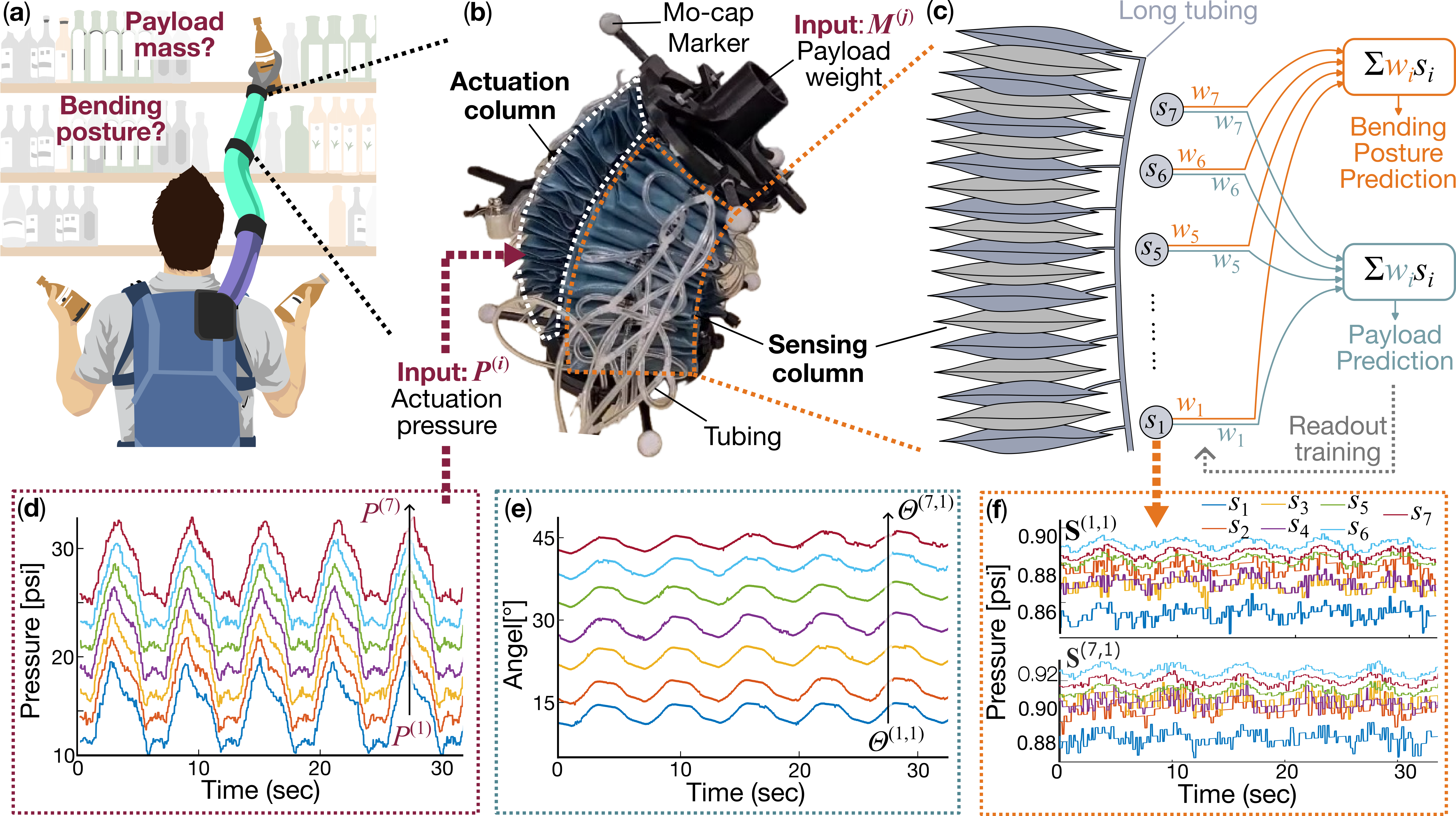}
    \caption{Setting up the robotic arm as a physical reservoir computer. (a-b) In this study, we use a fabric-based robotic arm segment as the physical testbed, which has three columns of pressurized pillows. (b-c) Working principle of the soft robotic arm reservoir. There are two different inputs to the reservoir: actuation pressure $P^{(i)}$ and end payloads $M^{(j)}$. Seven pressure sensors are distributed throughout the sensing column, and their readings (aka. the reservoir states) will be plugged into weighted linear summation for readout training and prediction. (d) Measured actuation pressures $s_{in}(t)$ sent to the robotic arm. (e) The corresponding robot bending angle captured by the motion capture system (without any payload). (f) Pressure sensor readings from actuation pressure input $P^{(1)}$ \textcolor{red}{and $P^{(7)}$ under} payload input $M^{(1)}$, these data constitute pressure state vector $\mathbf{S}^{(1,1)}(t)$ \textcolor{red}{and $\mathbf{S}^{(7,1)}(t)$}.}   
    \label{fig: setup}
\end{figure}

Therefore, in this study, we adopt a fabric-based, pneumatic soft robotic arm as the physical platform to examine the feasibility and performance of using simple pressure reading, combined with physical reservoir computing, to achieve proprioceptive and exteroceptive information perception tasks (Fig. \ref{fig: setup}). Here, pressure sensors are inherently available in the pneumatic system, so using them would not significantly increase the robot's complexity. Through extensive experimentation and readout training based on the physical reservoir computing framework, we uncovered that \textit{a simple, weighted linear summation of embedded pressure sensor readings can simultaneously offer us accurate predictions of bending posture (proprioceptive) and end payload mass (exteroceptive)} (Figs. \ref{fig: vision}, \ref{fig: setup}).  We also examined how to use the \textit{minimum} amount of pressure reading to obtain these predictions accurately. 

In what follows, the second section of this paper briefly describes the experimental setup of the fabric-based pneumatic robotic arm and the corresponding physical reservoir computing framework. The third section details the results for single and multi-task information perception. This paper ends with a conclusion and a discussion of future work.

\section{Setting up the Robotic Arm as Physical Reservoir Computer}

\subsection{Pneumatic Arm Test Platform}
We employed a fabric-based soft robot arm we had previously developed as the physical testing platform \cite{nguyen2019fabric}. This robotic arm consists of three parallel columns of ``pouches,'' each is a rectangular-shaped pneumatic pillow (Fig. \ref{fig: setup}a,b). The pouches in each column are connected in series via pneumatic tubing. Therefore, the soft robotic arm bends when the pouches on each column are inflated with different pressures and pushed against each other.

In this work, we inflate one of the three columns to generate a bending motion. This column is hence called the ``actuation column.'' Meanwhile, we inflate the second and half number of the third columns of pneumatic pouches with regulated and constant pressure to maintain the robotic arm's overall stiffness. The remaining pillows in the third column function as the sensing units. Therefore, we call this third column the ``sensing column'' (Fig. \ref{fig: setup}a,b).

In total, there are 10 air pouches for sensing, and they are distributed throughout the span of the robotic arm. These pouches are interconnected with a tubing network. A binary valve separates these sensing pouches and tubings from the external pressure supply, ensuring that a fixed amount of pressurized air remains sealed in them throughout each experiment. However, the sealed air can freely flow between these sensing pouches. \textcolor{red}{Thus, we can still see different pressure responses at these seven locations, as shown in figure \ref{fig: setup}f. }Finally, we attach seven pressure sensors to the tubing network (located at the $1^\text{st}$, $2^\text{nd}$, $3^\text{rd}$, $4^\text{th}$, $5^\text{th}$, $7^\text{th}$, and $9^\text{th}$ air pourch from the base, Fig. \ref{fig: setup}c). \emph{It is critical to point out that data from these sensors captures the proprioceptive and exteroceptive information} (explained further in Section \ref{sec: PRCintro}).

In each test run, we first place a payload on the robot's top, with seven options of 0, 100, 140, 160, 200, 240, or 300 grams. For clarity, we label these payloads as $M^{(j)}$ ($j=1, \ldots, 7$), so that $M^{(1)}=0$ and $M^{(7)}=300$. We then inflate the actuation column of air pouches to create dynamic bending motions using one of the seven pre-programmed input pressure profiles (labeled as $P^{(i)}$, $i=1\ldots, 7$). These pressure profiles have different magnitudes, but all consist of eight identical ramp cycles $u^{(i)}(t)$ defined by

\begin{align} \label{eq: input1}
    u^{(i)}(t) &=
	\begin{cases}
 	      u_\text{min}^{(i)} + r_\text{up}  t & t \in [0,T_\text{peak}),\\
 	      u_\text{max}^{(i)} - r_\text{down}  (t - T_\text{peak}) &  t \in [T_\text{peak},T),
        \end{cases}\\
    T_\text{peak}^{(i)}  &= \frac{u_\text{max}^{(i)}-u_\text{min}^{(i)}}{r_\text{up}},\\
    T^{(i)}         &= \frac{u_\text{max}^{(i)}-u_\text{min}^{(i)}}{r_\text{up}} + \frac{u_\text{max}^{(i)}-u_\text{min}^{(i)}}{r_\text{down}}.
\end{align}

Here, $u_\text{max}^{(i)}$ and $u_\text{min}^{(i)}$ is the maximum and minimum pressures for profile $P^{(i)}$, so seven different sets of $u_\text{max}$ and $u_\text{min}$ are used in total. $r_\text{up}$ and $r_\text{down}$ are the rate of ramp-up (or down) during inflation (or deflation), and $r_\text{up}=r_\text{down}=$ 5psi/sec for all tests. $T_\text{peak}^{(i)}$ and $T^{(i)}$ are the time duration of the inflation and one complete ramp cycle, respectively. The actual driving pressure sent to the robotic arm (labeled as $s_{in}(t)$) is summarized in Fig. \ref{fig: setup}(d). Note that input profiles with different pressure levels also change the soft robot arm's overall stiffness, further diversifying its dynamic responses --- a desirable feature for physical reservoir computing.

To monitor the robotic arm's operation condition and obtain data for the subsequent reservoir computer training, we use a 10-bit ADC board to convert the analog measurement from the seven pressure sensors in the sensing column (labeled as $s_1(t), s_2(t), \ldots s_7(t)$). We also use a motion capture system, with six OptiTrack Prime$^{x}$13 cameras (NaturalPoint, Inc., Corvallis, OR), to measure the actual bending angle of the robotic arm $\Theta(t)$ (examples shown in Fig. \ref{fig: setup}e). Altogether, there are 49 different input conditions (7 pressure profiles $\times$ 7 payload masses), labeled by bracketed superscript $^{(i)}$ and $^{(j)}$. Correspondingly, 49 groups of pressure readings are collected for the subsequent reservoir computing training discussed in the subsequent Section \ref{sec: PRCintro}. 

\subsection{Physical Reservoir Computing} \label{sec: PRCintro}

Soft robots have been identified as a promising candidate for physical reservoir computing with excellent information processing capability \cite{nakajima2013soft, kawase2021pneumatic, li2012behavior, nakajima2017muscular, horii2021physical, nakajima2018exploiting, bhovad2021physical}. Here, we further investigate the potential of harnessing such information processing power in proprioceptive and exteroceptive information perception tasks.  

Our fundamental assumption is that as the robotic arm deforms to different postures or carries different amounts of payload, its internal dynamics will change accordingly. Such changes will be captured by the pressure distribution throughout the sensing column. If we treat the robotic arm as a physical reservoir computer, we can extract helpful information embodied in the pressure dynamics. \emph{After appropriate training, simple weighted linear summations of the pressure sensor data will yield accurate estimations of the robotic arm's (proprioceptive) bending angle and the (exteroceptive) payload weight}.

Figure \ref{fig: setup}(b, c) illustrates the reservoir computing framework. There are two types of input into the robotic arm reservoir: the actuation pressure $P^{(i)}$ defined in Eq. (\ref{eq: input1}) and the payload weight $M^{(j)}$. The pressurized robotic arm functions as the fixed \textit{reservoir kernel}. Data from the seven pressure sensors $\mathbf{S}(t)=[s_1(t), s_2(t), \ldots, s_7(t)]^\intercal$ constitute the \emph{pressure state vectors}.  Fig. \ref{fig: setup}(f) shows an example of the pressure state vector corresponding to the pressure profile $P^{(1)}(t)$ and $M^{(1)}$ (no end payload). For clarity, we label the pressure state vectors as $\textbf{S}^{(i,j)}(t)$, corresponding to input pressure profile $P^{(i)}$ and end payload $M^{(j)}$. Interestingly, the sensors near the arm tip ($s_5(t)$ to $s_7(t)$) give higher pressure readings and signal-to-noise ratio than other sensors. Nevertheless, all sensor data show significant common-signal-induced synchronization, necessary for considering a signal-driven dynamical system as a reservoir. To guarantee this condition applies, we discard the first 50 seconds of data (2000 frames with 40Hz sampling frequency) before training and testing to eliminate transient responses. Therefore, in each test, pressure input lasts 100 seconds, generating up to 1000 data points (from 50 to 75 seconds) for training and another 1000 samples (from 75 to 100 seconds) for prediction testing.

Training the physical reservoir involves finding a set of static readout weights $\mathbf{w}_\text{out}=[w_0, w_1, \ldots, w_7]^\intercal$ via linear regression such that
\begin{equation}
    \label{eq: LR}
      {\mathbf{w}_\text{out}}=\left[\mathbf{I} \; \mathbf{S}(t)\right]^{+}\hat{\mathbf{y}}(t)={\mathbf{\bar{\Phi}}(t)^{+}}\hat{\mathbf{y}}(t),
\end{equation}
where $\mathbf{I} $ is a column of ones for calculating the bias term $w_0$, ${{[\cdot]}^{+}}$ is the Moore–Penrose pseudo-inverse to accommodate non-square matrices, and $\hat{\mathbf{y}}(t)$ is the target output, which is defined according to the task. For example, if one wants to use the pressure readings to estimate the robotic arm's bending angle, $\hat{\mathbf{y}}(t)$ would be defined using the actual bending angle $\Theta(t)$ measured by the motion capture system. To estimate the payload, $\hat{\mathbf{y}}(t)$ would be defined based on the actual weight $M^{(j)}$. Moreover, one can assemble multiple target functions into a matrix $\hat{\mathbf{Y}}(t)$ for multi-tasking. This way, Eq. (\ref{eq: LR}) will output multiple sets to readout weights simultaneously. It is worth emphasizing that the following Section 3 will give a more detailed definition of $\hat{\mathbf{y}}(t)$ in different scenarios. 

Once the readout weights are obtained from training, the reservoir's \textit{predictions} are simply:

\begin{equation}
    \label{eq: output}
      y(t)=w_0+\sum\limits_{i=1}^7 w_i s_i(t).
\end{equation}

Note that these reservoir outputs are also time-series data, so the accuracy of these predictions is evaluated using the root mean squared error:

\begin{equation} \label{eq: RSME}
      e =\left[\frac{1}{Q}\sum_{k=1}^{Q}\left(y_k-\hat{y}_k\right)^2 \right]^{1/2},
\end{equation}
where $Q$ is the number of time steps in the testing period, $y_k$ is the reservoir output at the $j^\text{th}$ step, and $\hat{y}_k$ is the corresponding true value at the $k^\text{th}$ step, we refer this as the \textit{prediction error} hereafter.

Note that previous work on reservoir computing suggested that a preferred reservoir should combine linear and nonlinear responses so that it can possess memory capacity (a linear property) and process complex information (a nonlinear transformation) \cite{inubushi2017reservoir} . Such insights lead to two main questions about our study. First, will the proprioceptive and exteroceptive information perception tasks require different levels of nonlinearity? Second, if a perception task requires lesser nonlinearity, could we reduce training accordingly? To this end, we start with a single information perception task in the next section to identify the optimal readout training methods.

\section{Information Perception by Arm Reservoir}
\subsection{Proprioceptive Bending Posture Prediction}

This task explores how the robotic arm reservoir could utilize its pressure dynamics to predict its bending posture. Since this is the first portion of this study, we simplify the test setup by removing the end payload (aka. $M^{(1)}=0$) and applying all seven pressure profiles ($P^{(1)}$ to $P^{(7)}$), resulting in seven different input conditions. Naturally, one would involve all these seven input conditions for readout training and ensure accurate reservoir predictions. Indeed, this is the approach used by the previous studies of using physical reservoir computing for information perception \cite{yu2022tapered, caluwaerts2013locomotion, sumioka2021wearable, hayashi2022multi}. However, this approach might not be suitable for soft robots because they frequently encounter complex and rapidly changing working conditions beyond training. \textcolor{red}{Involving data from all working conditions in the readout training process can be computationally expensive and unrealistic. }Therefore, we examine how the robotic arm reservoir can work in conditions not present in the training set.

\begin{figure}[ht!]
    \centering
    \includegraphics[width=6.0in]{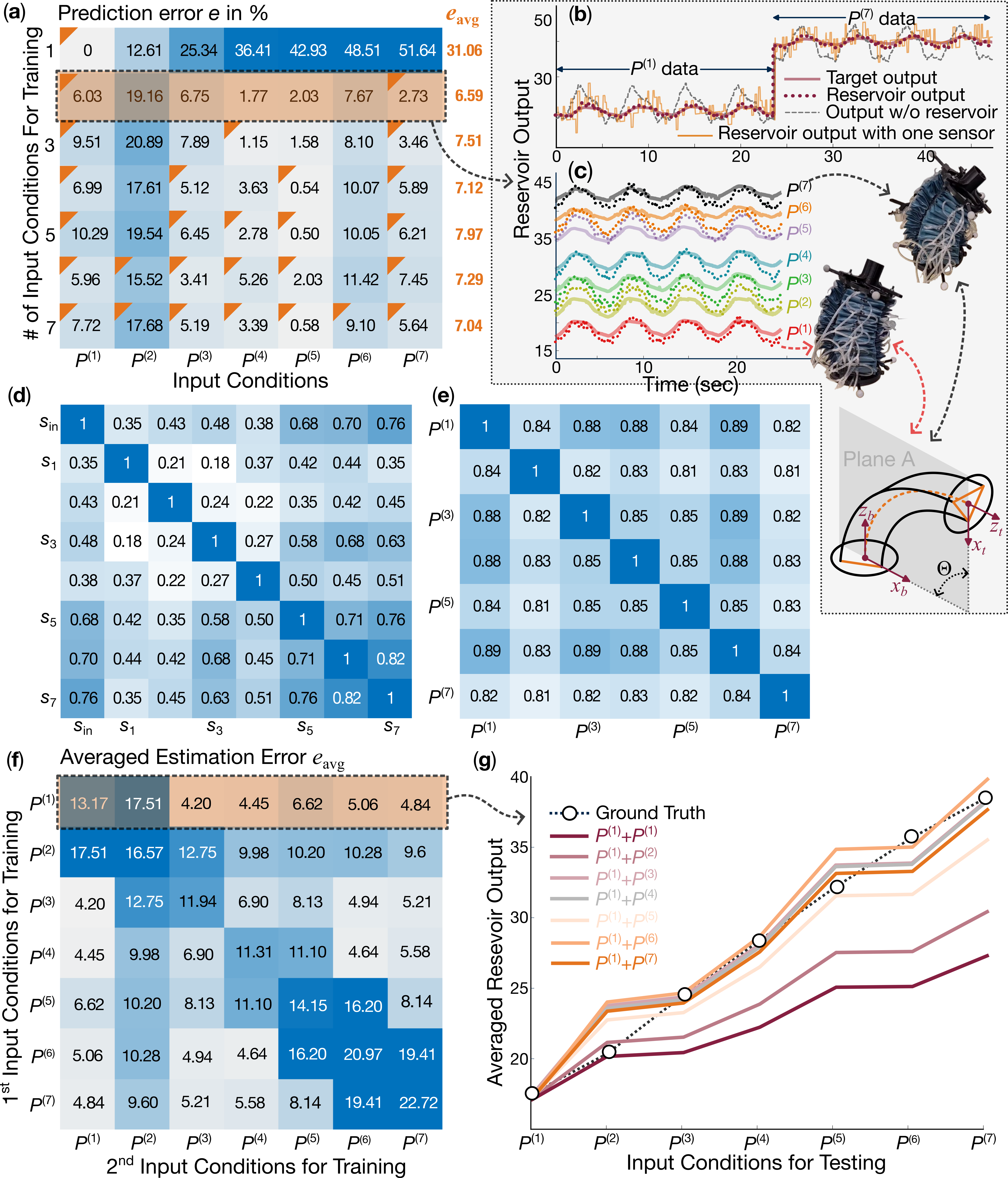}
    \caption{Predicting the bending angle under seven actuation pressure profiles $P^{(1)} \ldots P^{(7)}$ and no payloads. 
    (a) Prediction errors in the bending angle under each input pressure profile using different set-ups for readout training. Each row in this matrix corresponds to a unique combination of input conditions for readout training (marked by small triangles). For example, the second row shows the prediction errors from readout training with $P^{(1)}$ and $P^{(7)}$ data, and the last row shows the prediction errors with training data from all seven input conditions. Each column of this matrix shows the arm reservoir's prediction error corresponding to the seven actuation pressure profiles. 
    (b) Detailed training setup using two input conditions $P^{(1)}$ and $P^{(7)}$.
    (c) The corresponding predictions, where the solid lines are ground truth from the mo-cap system, and dashed lines are the reservoir's predictions.  
    (d) Correlation matrix for 7 different sensor readings under $P^{(1)}$. 
    (e) Correlation matrix of sensor data $s_7(t)$ under 7 different input conditions. 
    (f) The averaged prediction errors based on different choices of two input conditions for reading training.    
    (g) The corresponding prediction results under different pairs for training setup corresponding to the first row in (f).}
    \label{fig: bending}
\end{figure}

To this end, we conduct an extensive comparative analysis and discover that only two sets of training data from two different pressure inputs are sufficient to obtain accurate proprioceptive information perception. Figure \ref{fig: bending}(a) summarizes this analysis result. The first row of this colormap shows the reservoir's prediction errors if we involve only one input condition for training. That is, we only use the pressure reading $\mathbf{S}^{(1,1)}(t)$ from $P^{(1)}$ and the corresponding motion capture data as the targeted output ($\hat{\mathbf{y}}(t)=\Theta^{(1,1)}(t)$). One thousand data points from $\mathbf{S}^{(1,1)}(t)$ and $\hat{\mathbf{y}}(t)$ (from 50 to 75 seconds) are plugged into Eq. (\ref{eq: LR}) to obtain the readout weights $\mathbf{w}_\text{out}$. The training itself is a success: We use Eq. (\ref{eq: output}) to apply these readout weights to the testing portion of $\mathbf{S}^{(1,1)}(t)$, obtaining the reservoir's prediction $\mathbf{y}(t)$ (corresponding to 75 to 100 seconds of the experiment). The corresponding prediction error, defined by Eq. (\ref{eq: RSME}), is negligible and indicated by the colormap value at the top left corner (Fig. \ref{fig: bending}a). 

However, it is worth reminding that this set of readouts is obtained only by the data from input condition $P^{(1)}$. If we apply these readouts to the pressure readings from other input conditions $P^{(2)}$ to $P^{(7)}$, the corresponding reservoir predictions yield significant errors (the rest of Fig. \ref{fig: bending}(a)'s first row). For example, applying $\mathbf{w}_\text{out}$ to $\mathbf{S}^{(7,1)}(t)$ will yield a large 51.64\% rms error.

Therefore, we try a new set of readout weights to address this issue by involving the data from two input conditions for training. More specifically, we compile a larger pressure state matrix by simply assembling those from $P^{(1)}$ and $P^{(7)}$ so that $\mathbf{S}^*=\left[\mathbf{S}^{(1,1)};\mathbf{S}^{(7,1)}\right]$.  Similarly, a new targeted output is compiled so that $\hat{\mathbf{Y}}^*(t)= \left[\Theta^{(1,1)}(t);\Theta^{(7,1)}(t)\right]$ (Fig. \ref{fig: bending}b).  Then, we substitute these compiled and larger state vectors and targeted outputs into the reservoir training equation Eq. (\ref{eq: LR}) to obtain a new set of readout weights $\mathbf{w}_\text{out}^*$. Remarkably, one can apply these new readouts to pressure readings from all input conditions --- beyond the conditions used for training --- and still yield accurate bending angle predictions (see the second row of Fig. \ref{fig: bending}a and Fig. \ref{fig: bending}c).  

More interestingly, if we involve more than two input conditions in the training step, the robotic reservoir does not provide a more accurate prediction (as shown in rows 3 to 7 of Fig. \ref{fig: bending}a). Here, we define an ``Averaged Estimation Error $e_\text{avg}$'' as the averaged prediction errors corresponding to different training approaches. This way, $e_\text{avg}$ of training with only $P^{(1)}$ data is the averaged error of Fig. \ref{fig: bending}(a)'s first row: 31.06\%. $e_\text{avg}$ of training with $P^{(1)}$ and $P^{(7)}$ data is much lower at 6.59\% (second row). Comparing $e_\text{avg}$ of different training approaches reveals that training with two input conditions is even more accurate than training with all input conditions (although only by a small margin). 

One possible explanation is that the physical mapping between bending angle and arm pressure dynamics is relatively linear, so more training data might lead to overfitting. To this end, we calculate the correlation matrix between sensor readings in the same test (Fig. \ref{fig: bending}d) and the readings from the $s_7$ sensor between different input conditions (Fig. \ref{fig: bending}e). Overall, the data under various scenarios show relatively high correlations. 

Naturally, the next question arises: If combining two input conditions yields the best reservoir training results, which two input conditions should one use? To this end, we conduct another comprehensive analysis by comparing all possible combinations of two input conditions. That is, we compile the pressure sensor and motion capture data from two input conditions to perform readout training --- with the same setup explained above --- and then apply the readout to all input conditions to assess the averaged estimation error. The results are summarized in Fig. \ref{fig: bending}(f). Taking its first row as the example: One input condition used for training is always the lowest input pressure profile $P^{(1)}$, and the other one varies between $P^{(2)}$ to $P^{(7)}$. One can see that $e_\text{avg}$ is reasonably low except for training with $P^{(1)}$ or $P^{(2)}$, probably because sensors $s_1$ and $s_2$ are close to the robotic arm's fixed base and they can't capture rich pressure dynamics. Figure \ref{fig: bending}(g) compared the performance differences for each training pair with more detail.  

In summary, the optimized readout training approach for proprioceptive bending angle prediction is to use data from only two input conditions, and one of them should come from $P^{(3)}$ to $P^{(7)}$. Such a training approach only uses about 14\% of the overall data while maintaining a promising prediction accuracy below 10\% for all input conditions. 

It is worth mentioning that in parallel to the training using $\mathbf{S}^{(i,j)}$, we also conduct a separate training without involving the robotic arm reservoir --- by only using the actuation pressure reading $s_{in}(t)$ and the sensor reading at the robotic arm's base $s_1(t)$. \textcolor{red}{In other word, we compile data coming from a separate sensor at the pressure inlet (not embedded in the robotic arm) under $P^{(1)}$ and $P^{(7)}$.}This set of results, however, failed to predict the bending angle property as shown by the thin dashed line in Fig. \ref{fig: bending}(b), indicating the importance of the robotic reservoir's pressure dynamics. It is also worth mentioning that while linear mapping has a significant role in this proprioceptive bending angle prediction, it is no longer the case for exteroceptive payload predictions, as discussed in the next section \ref{sec: payload}. \textcolor{red}{To compare training without reservoir and with the simplest reservoir (one sensor), we compile data from closest to base $s_1(t)$ under $P^{(1)}$ and $P^{(7)}$, the yellow line in \ref{fig: bending}(b), which already gives a better results than training without reservoir.}

\subsection{Exteroceptive Payload Prediction}\label{sec: payload}
In this task, we explore if a simple, weighted linear summation of pressure reading can give us an estimation of the mass of end payload $M^{(j)}$. To this end, we find that this exteroceptive task has to be divided into two consecutive steps: payload detection and mass estimation. Indeed, such a 2-step setup suggests the multi-tasking capability of the robotic arm reservoir: One can use the same pressure state vector to train multiple sets of readout weights simultaneously for different tasks (discussed further in Sec. \ref{sec: multi}).   

\begin{figure}[t!]
    \centering
    \includegraphics[scale=1.0]{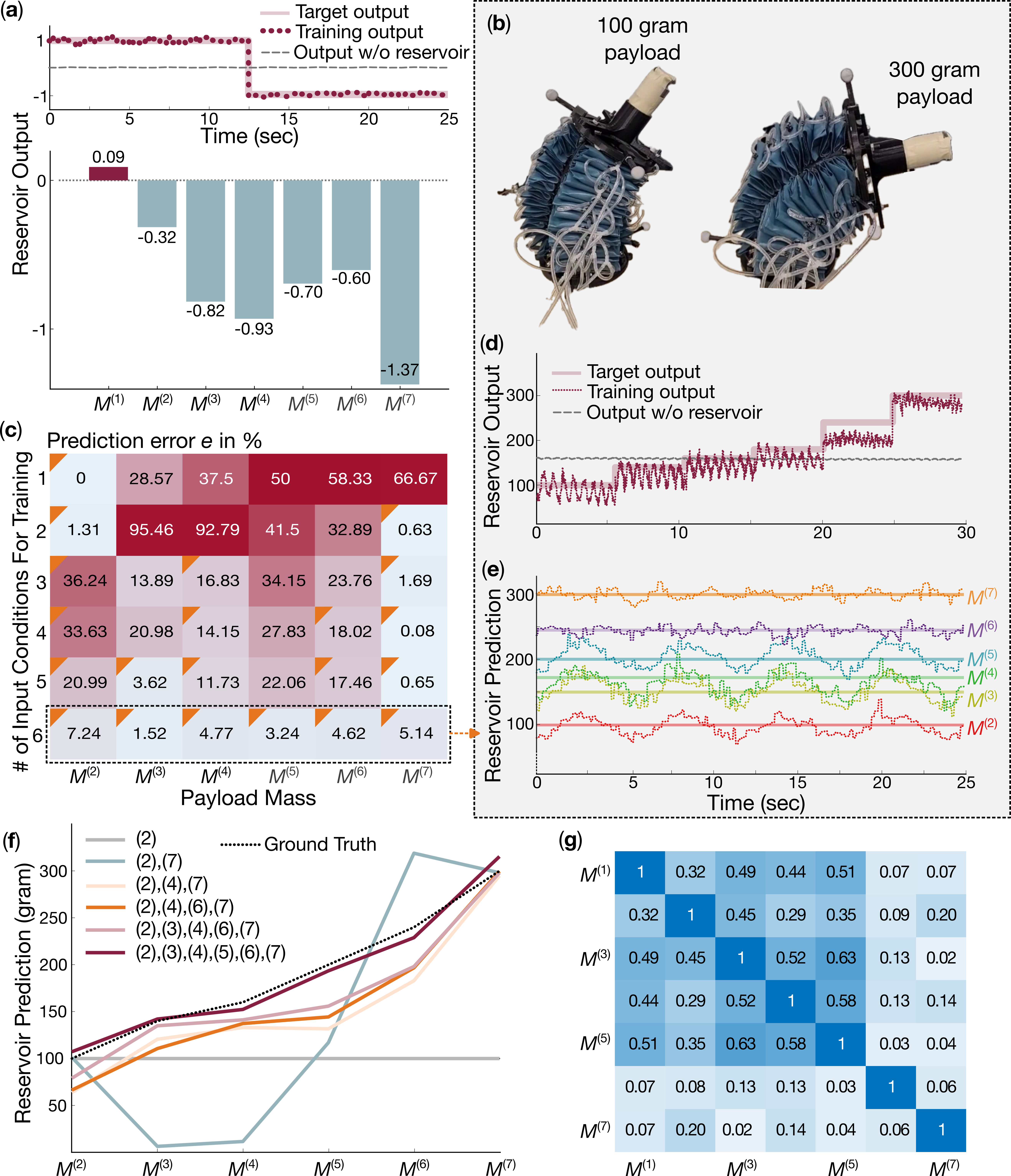}
    \caption{Predicting the existence and mass of the end payloads. 
    (a) Training and prediction results of the payload detection task to determine whether a payload exists. 
    (b) Pictures of the robotic arm's deformation under two payloads.
    (c) Prediction error (in \%) of payload mass based on different combinations of input conditions for readout training. Each row in this matrix corresponds to a unique selection of the input conditions (marked by the small triangles). For example, the second row shows the results from training with data from $m^{(2)}$ and $m^{(7)}$ inputs, and the last row with all six payloads. Each column of this matrix shows the arm reservoir's prediction errors of a particular payload.  
    (d) Detailed readout training setup and the training outputs using all non-zero payload mass input conditions. This setup corresponds to the last row in the colormap. 
    (e) The testing data corresponding to the best readout training setup.  Here, the thick solid lines are the true payload mass, and the thin lines are the reservoir's prediction.
    (f) The reservoir's prediction for the 6 different payloads using different readout training setups. This is the same data as shown in the colormap but presented in a different way.
    (g) Correlation matrix of sensor data $s_7(t)$ under 7 different payloads.}
    \label{fig: payload}
\end{figure}

The first step of payload detection aims to determine whether any payloads are present. To this end, we assemble the pressure sensor readings corresponding to no payload and most miniature payloads of 100 grams, in that $\mathbf{S}^{**}=[\mathbf{S}^{(1,1)};\mathbf{S}^{(1,2)}]$ (each containing 5 seconds of data), and the corresponding targeted output for training is 1 for no payload and -1 for 100 grams payload so that

\begin{equation}
    \label{eq:detect_target}
    {{\hat{\mathbf{y}}}_{2}}(t)=
    \begin{cases}
      1&(0<t<5 \text{ sec}),\\
      -1&(5 \le t<10 \text{ sec}).\\ 
    \end{cases}
\end{equation}

Figure \ref{fig: payload}(a) illustrates this training. We then apply the readouts obtained from this training to all the pressure state vectors $\mathbf{S}^{(1,j)}$, $j=1,\ldots,7$. Our idea is that an averaged positive reservoir output indicates no payload and a negative output indicates that payload exists. The arm reservoir succeeds in this step (Fig. \ref{fig: payload}a). 

We also conducted parallel training without involving the arm reservoir, but this readout failed to provide any successful results, further elucidating the need to involve the pressure dynamics for physical reservoir training. 

The second step of the payload prediction task aims to directly predict its mass by a linear summation of pressure readings. To this end, we recorded the pressure readings from six different input conditions. That is, the payload weight varies from $M^{(2)}$ to $M^{(7)}$, while the input pressure profile remains at the lowest level $P^{(1)}$.  Similar to the proprioceptive bending posture task, we explore how many input conditions are necessary for the readout training to predict payload mass accurately. To this end, we conduct another comparative analysis with different input conditions, summarized in Figure \ref{fig: payload}(c). Different rows of this colormap represent different combinations of input conditions used for readout training, as highlighted by the small triangles at the upper left corners. Take the second row as an example: This row results from using two input conditions for readout training. We compile 12.5 seconds of sensor reading corresponding to $M^{(2)}=100$ grams and $M^{(7)}=300$ grams so that $\mathbf{S}^{**}=[\mathbf{S}^{(1,2)};\mathbf{S}^{(1,7)}]$, and targeted output is a step function according to the actual payload so that,

\begin{equation}
    \label{eq:task 2 target}
    \mathbf{\hat{y}}_{2}(t)=
    \begin{cases}
      {{M}^{(2)}=100}&(0<t<12.5 \text{ sec}),\\
      {{M}^{(7)}=300}&(12.5\le t<25 \text{ sec}).\\ 
    \end{cases}
\end{equation}

Plugging $\mathbf{S}^{**}(t)$ and this $\hat{\mathbf{y}}_2(t)$ into the training Eq. (\ref{eq: LR}) will yield a set of readout $\mathbf{w}_\text{out}^{**}$.  Then, this readout can be applied to all the pressure readings $\mathbf{S}^{(1,j)}, j=2,\ldots,7$, using Eq. (\ref{eq: output}), for payload mass prediction. The corresponding prediction errors are reflected by the colormap.

However, unlike bending angle perception training, using two input conditions for readout training is not sufficient for payload mass estimation (Fig. \ref{fig: payload}c). The average estimation error decreases significantly as more input conditions are employed in readout training until all six payload inputs are involved and the average estimation error drops below 10\%. In this best-case setup, we compile 5 seconds of pressure data from all input conditions so that $\mathbf{S}^+=[\mathbf{S}^{(1,2)}; \mathbf{S}^{(1,3)}; \cdots \mathbf{S}^{(1,7)}]$. The corresponding targeted output $\mathbf{\hat{Y}}(t)$ becomes a 6-step function correspondingly.  Figure \ref{fig: payload}(d-f) shows this case's training and prediction results. \textcolor{red}{The oscillatory pattern could still be observed in reservoir output because of the limited nonlinear dynamics that our physical reservoir could provide. If we could modify the robotics arm design to generate more diverse dynamics, the oscillatory pattern could be decreased.Meanwhile, the oscillating nature becomes relatively minor when the payload is large (as shown in Figure \ref{fig: payload}e). }

Looking into the correlation matrix of sensor $s_7(t)$ data between different input payload conditions (Fig. \ref{fig: payload}g), one can see most correlation values are closer to zero, indicating a strong nonlinearity. This nonlinear mapping between payload inputs and pressure dynamics might explain the demanding requirements for readout training. In other words, the nonlinear body dynamics are critical in mapping pressure sensor data into payload mass predictions.

\subsection{Sensor Data Analysis}
Once we select the optimal input conditions for readout training (for proprioceptive bending posture prediction or exteroceptive payload prediction), two more questions remain: 1) How many training data samples are sufficient for each readout training? 2) How many pressure sensors are required to complete a successful readout training? Answers to these questions are crucial for reducing the training cost.

\begin{figure}[ht!]
    \centering
    \includegraphics[]{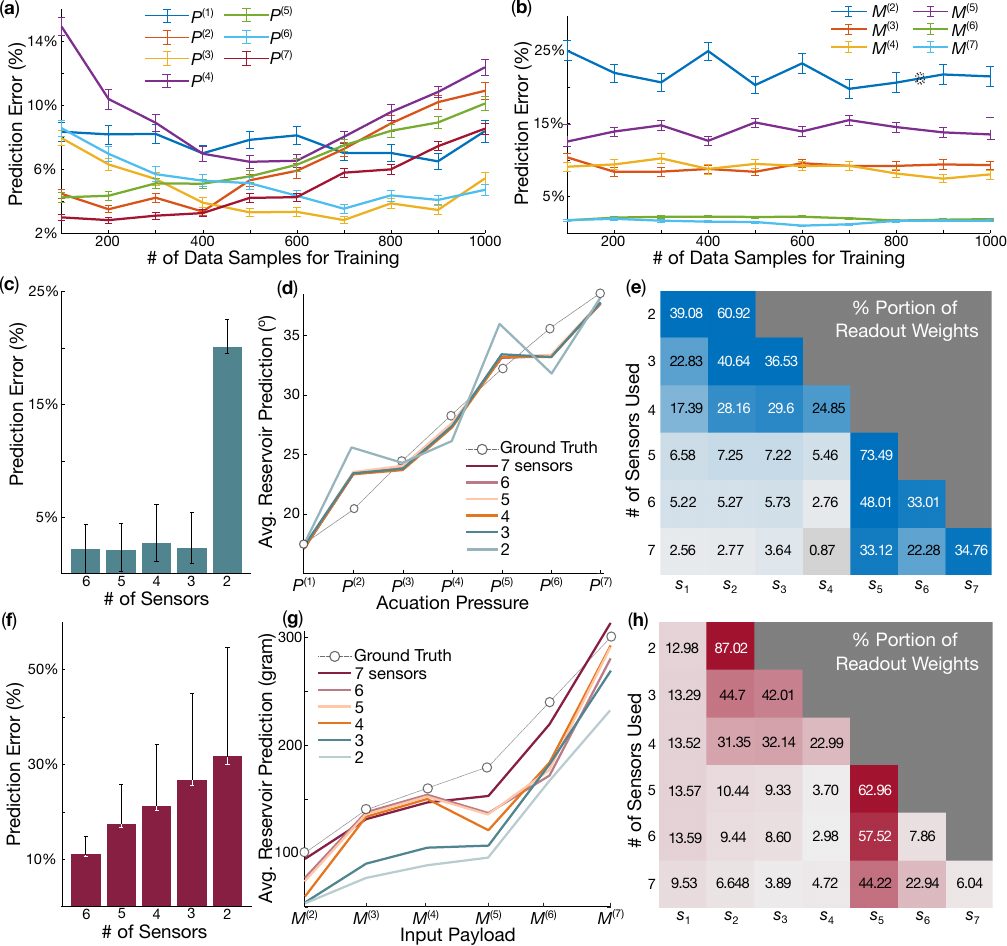}
    \caption{Arm reservoir's prediction accuracy with reduced training data or reduced number of sensors. 
    (a) Averaged prediction errors for bending posture tasks using different amounts of data samples for readout training. The seven lines correspond to the 7 different actuation pressure inputs. 
    (b) Averaged error for payload mass prediction using different amounts of data samples for readout training. The six lines represent the six non-zero input masses. 
    (c) The overall prediction errors with reduced numbers of sensors for training and prediction. The number of sensors decreases from 6 to 2 (left to right). 
    (d) Sample results of bending angle prediction with fewer sensors. 
    (e) The corresponding percentage portion of readout weights for each sensor involved. Each row adds up to 100\%.
    (e-h) Similar studies on reduced sensors for the payload mass prediction task.}
    \label{fig: data}
\end{figure}

To answer the first question, we keep the optimal selections of input conditions identified in the previous sections --- $P^{(1)}$ and $P^{(7)}$ for bending posture prediction and $M^{(2)}$ through $M^{(7)}$ for payload prediction --- and compared the results from using different amount of training data.  For example, for bending prediction, we conduct seven readout training using as few as 100 data samples (2.5 seconds of data) and as many as 1000 samples (25 seconds). Each training is repeated 10 times to ensure consistent outputs.  Meanwhile, the testing data is fixed at 25 seconds, and the corresponding prediction errors are compared in Figure \ref{fig: data}(a). Intuitively speaking, one would believe the more training data we use, the better the results will be. However, the results show that the 400 data samples (or 10 seconds of data) yield the best results, with all prediction errors falling below 10\%.  The prediction error actually increases when more than 500 data samples are used for readout training. On the other hand, the payload mass prediction errors stay roughly the same regardless of the amount of training data (Fig. \ref{fig: data}b). Main contribution to this results might be the more comprehensive training case set in payload task, making the prediction error less sensitive to the length of training data set.

So far, the training and testing involve data from all pressure sensors in the sensing column. What if we use fewer pressure sensors? We conduct a new training set using different sensor selections.

For proprioceptive bending posture prediction, we discover that three sensors are sufficient to obtain good performance, with errors within 5\%.  Figure \ref{fig: data}(c,d) summarizes the results of the bending angle prediction, and Figure \ref{fig: data}(e) details the readout weights corresponding to each sensor used for the proprioception task. Regardless of the sensors used, the sensor data closer to tip have the most significant influence (with the largest readout weight). This might be because the pressure dynamics near the tip can capture sufficient information about the bending kinematics, while pressure near the arm’s base is closer to input pressure. 

Exteroceptive payload prediction needs relatively more pressure sensor data. Figure \ref{fig: data}(f-h) shows that the average error progressively increases with fewer sensors, and at least six out of the seven sensors are needed to reduce the prediction error to below 10\%. In this case, sensors near the arm’s base contribute more than the proprioception task, meaning that the pressure dynamics throughout the robotic arm’s body contain useful information about the payload mass. Regardless, sensors near the tip still have more significant roles.

\subsection{Multi-Tasking} \label{sec: multi}

Physical reservoirs have demonstrated multi-tasking ability with emulation tasks \cite{hauser2011towards} (i.e., emulating several nonlinear filters with the same reservoir state vectors). Since the soft robot reservoir can project all input streams into a high-dimensional state space, it should also be able to extract two pieces of information from one set of pressure readings. We validate this capability by requiring the robotic arm reservoir to predict the body posture and payload mass simultaneously. 

\begin{figure}[ht!]
    \centering
    \includegraphics[scale=0.95]{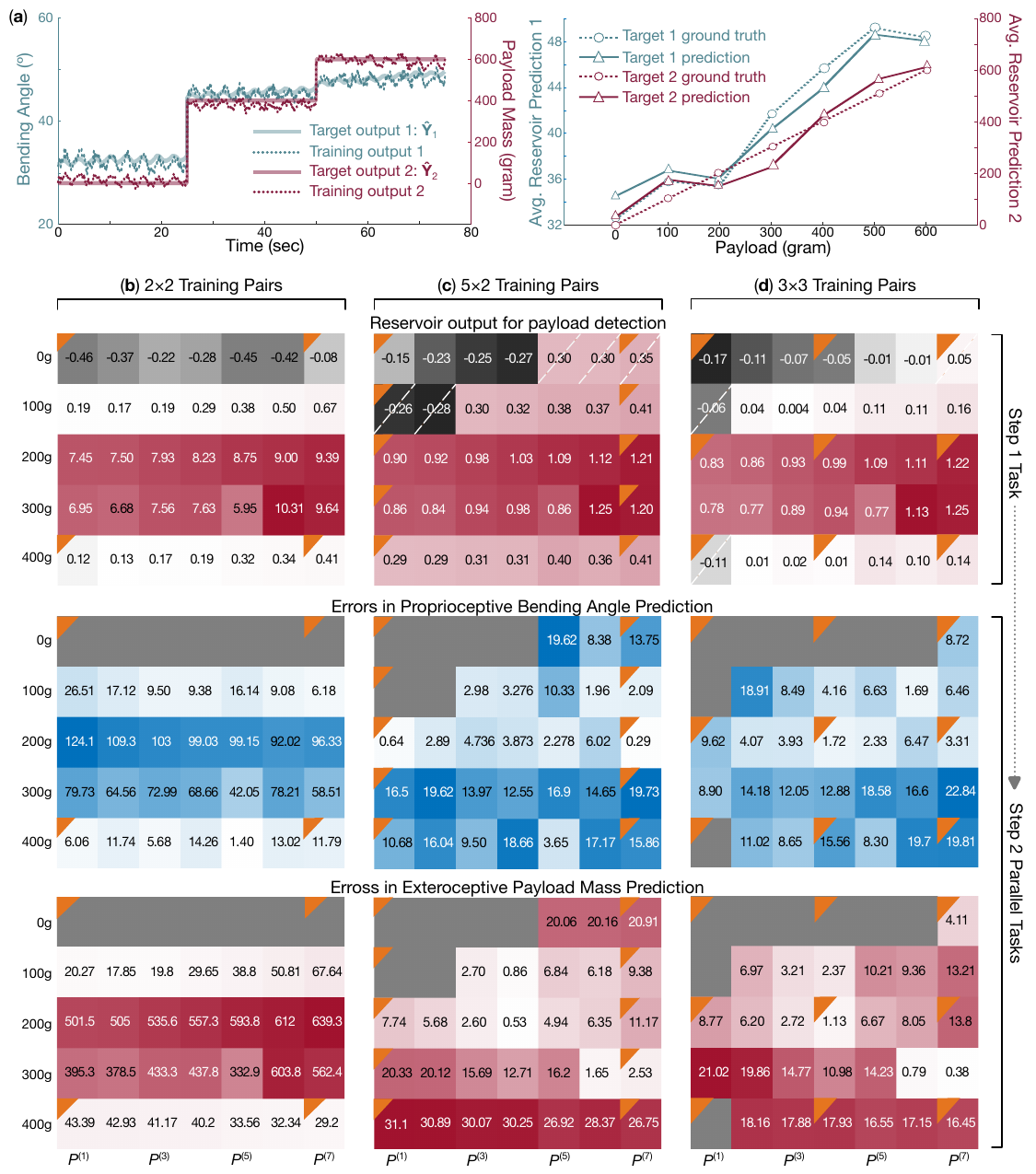}
    \caption{Multi-tasking with pneumatic arm reservoir. 
    (a) Training (left) and testing (right) results for simultaneous bending posture and payload mass prediction tasks. Target 1 refers to bending angle prediction, and target 2 is payload mass estimation. 
    (b) Multi-tasking performance with complex and realistic input conditions. Here, the column of three colormaps summarizes the results of three tasks. We divide the three tasks into two consecutive steps. The first step is payload detection; the second step includes bending angle and payload mass prediction. The four small triangles in each colormap indicate the input conditions used for readout training.
    (c) Another study using 10 input conditions for readout training, as shown by the small triangles. The prediction errors in the second step are much smaller, but the reservoir failed in payload detection in some cases, highlighted by the white diagonal lines.
    (d) Another study using 9 input conditions for readout training, this case yielded the best overall results. }
    \label{fig:multi-task}
\end{figure}

We consider two case studies for multi-tasking. In the first case, the payload varies from 0, 100, 200, 300, 400, 500 to 600 grams, while the input pressure is fixed at $P^{(7)}$.  (Notice these payloads are different from those in the previous sections, so we move away from the $M^{(j)}$ notations to avoid confusion.) Due to the payload difference, the robotic arm’s bending posture varies, even though the actuation pressure does not change. We compile pressure sensor readings corresponding to 0-, 400-, and 600-gram payload inputs for readout training. Meanwhile, two data sets are compiled as the targeted output for readout training. One is the bending angle obtained from the motion tracking system ($\mathbf{\hat{Y}}_1$ in Fig. \ref{fig:multi-task}a). The other is the payload mass $\mathbf{\hat{Y}}_2$ defined as a piece-wise step function. By assembling these two targeted outputs into one matrix $\mathbf{\hat{Y}} = [\mathbf{\hat{Y}}_1; \mathbf{\hat{Y}}_2]$ and substituting it into to Eq. (\ref{eq: LR}), we obtain two groups of readout weights, one for bending posture prediction and the other for payload mass prediction.  Then, these two groups of readout weights are applied to the pressure readings from all seven input conditions to simultaneously predict the bending angle and payload mass. Figure \ref{fig:multi-task}(a) (right) summarizes the prediction results, showing that the reservoir's outputs match the ground truth well.

The success of this first case study shows that multi-tasking under changing payload could be accomplished with the same training approach as the single-task information perception. However, if the input becomes more realistic and complex---both actuation pressure and payload change simultaneously---the readout training needs to be conducted more carefully and comprehensively. To this end, we obtained a new data set by combining the seven actuation pressure profiles ($P^{(1)}$ to $P^{(7)}$) and five payloads (0, 100, 200, 300, and 400 grams), so there are altogether 35 sets of pressure sensor readings. We then examine whether the robotic reservoir can simultaneously (1) predict the bending gesture, (2) detect the payload's existence, and (3) predict the payload mass if it exists. 
Figure \ref{fig:multi-task}(b-d) summarizes the multi-tasking performance based on three different readout training setups. 

For example, in the first column of colormaps in Fig. \ref{fig:multi-task}(b), we assemble ($2 \times 2=$)4 sets of pressure sensor readings from 4 different input conditions: $P^{(1)}$ with no payload, $P^{(7)}$ with no payload, $P^{(1)}$ with the 400-gram payload, and $P^{(7)}$ with 400-gram payload (as indicated by the small triangles).  For the targeted output for readout training, we assemble three data sets corresponding to the three tasks: the bending angle from the motion capture system, a ``-1 or 1'' step function indicating the existence of payloads (similar to Eq. (\ref{eq:detect_target})), and another step function corresponding to the real payload mass (similar to Eq. (\ref{eq:task 2 target})).  Plugging these compiled pressure data and target functions into Eq. (\ref{eq: LR}) yields three groups of readouts. Then, we applied these readouts to all the pressure data from the other 31 input conditions to assess the performance of the three prediction tasks mentioned above. Therefore, the first column of three colormaps in Fig. \ref{fig:multi-task}(b) details the reservoir's performance for each task, respectively.

More specifically, we divide the three tasks into two steps. The first step is payload detection tasks. If the payload exists (i.e. the reservoir's prediction output is larger than zero), the robotic arm reservoir will proceed to the next step of two other prediction tasks. Looking at the first colormap in Fig. \ref{fig:multi-task}(b), one can see that the robotic reservoir performed perfectly in payload detection.  It gives negative outputs when the payload is absent (first row) and positive values otherwise (second to fifth rows). However, it performed poorly in the second step of bending angle and payload mass prediction tasks, yielding large prediction errors in some cases, as indicated in the second and third colormaps in the first column (Fig. \ref{fig:multi-task}b).

Therefore, we experiment with two more readout training setups involving data from more input conditions. For example, the second column of colormaps  (Fig. 
\ref{fig:multi-task}c) summarizes the results from training with ($5 \times 2=$) 10 input conditions, and the third column (Fig. \ref{fig:multi-task}d) involves ($3 \times 3=$) 9  input conditions, all indicated by the small triangles.  One can see a significant drop in the second step's prediction errors, especially with training by $3 \times 3$ input conditions. On the other hand, the first step of payload detection starts to show errors.  

Clearly, there is a trade-off between different tasks. More interestingly, we tried adding more input conditions into readout training, but it did not generate better performance in any tasks. This indicates that we already reached the performance limit of the robotic reservoir for multitasking. To further increase the performance, we could either refine the robot's design to generate more diversified dynamics (e.g., non-uniform sensor distribution) or add more pressure sensors into the arm. Regardless, these results clearly elucidate the potential of robotic reservoirs for multi-functional sensing and the most effective approach towards readout training.

\section{Summary and Discussion}
We experimentally demonstrate the proprioceptive and exteroceptive information perception capabilities embodied in a fabric-based soft robotic arm using the physical reservoir computing framework --- A simple, weighted linear summation of the readings from embedded pressure sensors can yield accurate predictions of the robotic arm's bending angle (proprioceptive) and payload mass (exteroceptive), either separately or simultaneously. However, careful readout training is necessary to guarantee acceptable prediction errors. We use comprehensive comparative analysis to show that data from two different input conditions are needed for proprioceptive bending prediction training. On the other hand, at least five input conditions are needed for exteroceptive payload prediction training. A more careful examination of the correlation between pressure sensor data reveals that nonlinear body dynamics are more critical in mapping from pressure data in payload information, while linear mapping is sufficient for bending perception. It is such a combination of linear and nonlinear body dynamics that makes the fabric-based robotic arm a capable reservoir kernel to achieve proprioceptive and exteroceptive information perceptions. \textcolor{red}{It shoud be noted that we are only using a "standard" pneumatically driven soft robot, and probably because of this choice, the final results might seem limited in some scenarios. However, we show that by carefully designing the readout training setup, we only need to use a subset of operating conditions for successful output. In other word,even when we are working with a robot that was not initially designed for PRC, we can still harness useful computing power through careful sensing and advanced readout training, indicating this method could broadly impact the soft robotics discipline.}

In summary, the results of this paper elucidate the exciting opportunity of leveraging physical reservoir computing to extract meaningful information for soft robotic control with only pressure sensors, relieving us from using complex and specialized sensors. The comprehensive study to identify the optimized readout training approach for different tasks not only shows the importance of combined linearity and nonlinearity in body dynamics for reservoir kernels. One can also extend these readout training approaches to other soft robotic systems for similar information perception tasks. In future work, we will consider optimizing the structure design, sensor distribution, and tube arrangement to increase the inherent computing capacity of the robot reservoir. This way, more sophisticated information perception tasks, such as predicting the robotic body's stiffness or detecting collision with obstacles, might be achievable. This information will be valuable for more effective and intelligent feedback control for soft robots.

\medskip

\medskip


\medskip
\textbf{Acknowledgements} \par 
The authors acknowledge the support from the National Science Foundation (CMMI-1933124, CMMI-2328522, CMMI-1800940, and EFRI-2422340).
\medskip

%
\bibliographystyle{MSP}
\bibliography{reference,reference_Zhang}

\begin{thebibliography}{10}
\providecommand{\url}[1]{\texttt{#1}}
\providecommand{\urlprefix}{URL }

\bibitem{laschi2016soft}
C.~Laschi, B.~Mazzolai, M.~Cianchetti,
\newblock \emph{Science robotics} \textbf{2016}, \emph{1}, 1 eaah3690.

\bibitem{polygerinos2017soft}
P.~Polygerinos, N.~Correll, S.~A. Morin, B.~Mosadegh, C.~D. Onal, K.~Petersen, M.~Cianchetti, M.~T. Tolley, R.~F. Shepherd,
\newblock \emph{Advanced Engineering Materials} \textbf{2017}, \emph{19}, 12 1700016.

\bibitem{yumbla2021human}
E.~Q. Yumbla, Z.~Qiao, W.~Tao, W.~Zhang,
\newblock \emph{Current Robotics Reports} \textbf{2021}, 1--15.

\bibitem{pinskier2022bioinspiration}
J.~Pinskier, D.~Howard,
\newblock \emph{Advanced Intelligent Systems} \textbf{2022}, \emph{4}, 1 2100086.

\bibitem{zaidi2021actuation}
S.~Zaidi, M.~Maselli, C.~Laschi, M.~Cianchetti,
\newblock \emph{Current Robotics Reports} \textbf{2021}, \emph{2}, 3 355.

\bibitem{della2023model}
C.~Della~Santina, C.~Duriez, D.~Rus,
\newblock \emph{IEEE Control Systems Magazine} \textbf{2023}, \emph{43}, 3 30.

\bibitem{wang2022control}
J.~Wang, A.~Chortos,
\newblock \emph{Advanced Intelligent Systems} \textbf{2022}, \emph{4}, 5 2100165.

\bibitem{hegde2023sensing}
C.~Hegde, J.~Su, J.~M.~R. Tan, K.~He, X.~Chen, S.~Magdassi,
\newblock \emph{ACS nano} \textbf{2023}, \emph{17}, 16 15277.

\bibitem{faris2023proprioception}
O.~Faris, R.~Muthusamy, F.~Renda, I.~Hussain, D.~Gan, L.~Seneviratne, Y.~Zweiri,
\newblock \emph{Soft Robotics} \textbf{2023}, \emph{10}, 3 467.

\bibitem{shu2023machine}
S.~Shu, Z.~Wang, P.~Chen, J.~Zhong, W.~Tang, Z.~L. Wang,
\newblock \emph{Advanced Materials} \textbf{2023}, \emph{35}, 18 2211385.

\bibitem{dickey2008eutectic}
M.~D. Dickey, R.~C. Chiechi, R.~J. Larsen, E.~A. Weiss, D.~A. Weitz, G.~M. Whitesides,
\newblock \emph{Advanced functional materials} \textbf{2008}, \emph{18}, 7 1097.

\bibitem{wang2018self}
T.~Wang, Y.~Zhang, Q.~Liu, W.~Cheng, X.~Wang, L.~Pan, B.~Xu, H.~Xu,
\newblock \emph{Advanced Functional Materials} \textbf{2018}, \emph{28}, 7 1705551.

\bibitem{atalay2017batch}
A.~Atalay, V.~Sanchez, O.~Atalay, D.~M. Vogt, F.~Haufe, R.~J. Wood, C.~J. Walsh,
\newblock \emph{Advanced Materials Technologies} \textbf{2017}, \emph{2}, 9 1700136.

\bibitem{larson2016highly}
C.~Larson, B.~Peele, S.~Li, S.~Robinson, M.~Totaro, L.~Beccai, B.~Mazzolai, R.~Shepherd,
\newblock \emph{science} \textbf{2016}, \emph{351}, 6277 1071.

\bibitem{kim2021probabilistic}
D.~Kim, M.~Park, Y.-L. Park,
\newblock \emph{IEEE Transactions on Robotics} \textbf{2021}, \emph{37}, 5 1728.

\bibitem{loo2022robust}
J.~Y. Loo, Z.~Y. Ding, V.~M. Baskaran, S.~G. Nurzaman, C.~P. Tan,
\newblock \emph{Soft Robotics} \textbf{2022}, \emph{9}, 3 591.

\bibitem{dai2021flexible}
Y.~Dai, S.~Gao,
\newblock \emph{IEEE Sensors Journal} \textbf{2021}, \emph{21}, 23 26355.

\bibitem{scimeca2019model}
L.~Scimeca, J.~Hughes, P.~Maiolino, F.~Iida,
\newblock \emph{IEEE Robotics and Automation Letters} \textbf{2019}, \emph{4}, 3 2479.

\bibitem{nguyen2019soft}
P.~H. Nguyen, C.~Sparks, S.~G. Nuthi, N.~M. Vale, P.~Polygerinos,
\newblock \emph{Soft robotics} \textbf{2019}, \emph{6}, 1 38.

\bibitem{nakajima2013soft}
K.~Nakajima, H.~Hauser, R.~Kang, E.~Guglielmino, D.~G. Caldwell, R.~Pfeifer,
\newblock \emph{Frontiers in computational neuroscience} \textbf{2013}, \emph{7} 91.

\bibitem{tanaka2019recent}
G.~Tanaka, T.~Yamane, J.~B. H{\'e}roux, R.~Nakane, N.~Kanazawa, S.~Takeda, H.~Numata, D.~Nakano, A.~Hirose,
\newblock \emph{Neural Networks} \textbf{2019}, \emph{115} 100.

\bibitem{nakajima2020physical}
K.~Nakajima,
\newblock \emph{Japanese Journal of Applied Physics} \textbf{2020}, \emph{59}, 6 060501.

\bibitem{pieters2022leveraging}
O.~Pieters, T.~De~Swaef, M.~Stock, F.~Wyffels,
\newblock \emph{Scientific Reports} \textbf{2022}, \emph{12}, 1 12594.

\bibitem{shougat2021hopf}
M.~R. E.~U. Shougat, X.~Li, T.~Mollik, E.~Perkins,
\newblock \emph{Scientific Reports} \textbf{2021}, \emph{11}, 1 19465.

\bibitem{tsunegi2019physical}
S.~Tsunegi, T.~Taniguchi, K.~Nakajima, S.~Miwa, K.~Yakushiji, A.~Fukushima, S.~Yuasa, H.~Kubota,
\newblock \emph{Applied Physics Letters} \textbf{2019}, \emph{114}, 16 164101.

\bibitem{bhovad2021physical}
P.~Bhovad, S.~Li,
\newblock \emph{Scientific Reports} \textbf{2021}, \emph{11}, 1 1.

\bibitem{inubushi2017reservoir}
M.~Inubushi, K.~Yoshimura,
\newblock \emph{Scientific reports} \textbf{2017}, \emph{7}, 1 1.

\bibitem{wang2023building}
J.~Wang, S.~Li,
\newblock \emph{Advanced Intelligent Systems} \textbf{2023}, \emph{5}, 9 2300086.

\bibitem{kawase2021pneumatic}
T.~Kawase, T.~Miyazaki, T.~Kanno, K.~Tadano, Y.~Nakajima, K.~Kawashima,
\newblock \emph{Sensors and Materials} \textbf{2021}, \emph{33}, 8 2803.

\bibitem{nguyen2019fabric}
P.~H. Nguyen, I.~I. Mohd, C.~Sparks, F.~L. Arellano, W.~Zhang, P.~Polygerinos,
\newblock In \emph{2019 International Conference on Robotics and Automation (ICRA)}. IEEE, \textbf{2019} 8429--8435.

\bibitem{li2012behavior}
T.~Li, K.~Nakajima, M.~Cianchetti, C.~Laschi, R.~Pfeifer,
\newblock In \emph{2012 IEEE International Conference on Robotics and Automation}. IEEE, \textbf{2012} 4918--4924.

\bibitem{nakajima2017muscular}
K.~Nakajima,
\newblock In \emph{Brain Evolution by Design}, 403--414. Springer, \textbf{2017}.

\bibitem{horii2021physical}
Y.~Horii, K.~Inoue, S.~Nishikawa, K.~Nakajima, R.~Niiyama, Y.~Kuniyoshi,
\newblock In \emph{ALIFE 2021: The 2021 Conference on Artificial Life}. MIT Press, \textbf{2021} .

\bibitem{nakajima2018exploiting}
K.~Nakajima, H.~Hauser, T.~Li, R.~Pfeifer,
\newblock \emph{Soft robotics} \textbf{2018}, \emph{5}, 3 339.

\bibitem{yu2022tapered}
Z.~Yu, S.~Perera, H.~Hauser, P.~R. Childs, T.~Nanayakkara,
\newblock \emph{IEEE Robotics and Automation Letters} \textbf{2022}, \emph{7}, 2 3608.

\bibitem{caluwaerts2013locomotion}
K.~Caluwaerts, M.~D'Haene, D.~Verstraeten, B.~Schrauwen,
\newblock \emph{Artificial life} \textbf{2013}, \emph{19}, 1 35.

\bibitem{sumioka2021wearable}
H.~Sumioka, K.~Nakajima, K.~Sakai, T.~Minato, M.~Shiomi,
\newblock In \emph{2021 IEEE/RSJ International Conference on Intelligent Robots and Systems (IROS)}. IEEE, \textbf{2021} 9504--9511.

\bibitem{hayashi2022multi}
H.~Hayashi, T.~Kawase, T.~Miyazaki, M.~Sogabe, Y.~Nakajima, K.~Kawashima,
\newblock In \emph{2022 IEEE/SICE International Symposium on System Integration (SII)}. IEEE, \textbf{2022} 578--584.

\bibitem{hauser2011towards}
H.~Hauser, A.~J. Ijspeert, R.~M. F{\"u}chslin, R.~Pfeifer, W.~Maass,
\newblock \emph{Biological cybernetics} \textbf{2011}, \emph{105}, 5 355.

\end{thebibliography}

\end{document}